\documentclass{article}

\usepackage{microtype}
\usepackage{subfigure}
\usepackage{booktabs} % for professional tables

\usepackage{hyperref}

\usepackage[accepted]{icml2018}
\usepackage{wrapfig}
\usepackage{amsfonts}
\usepackage{amsmath}
\usepackage{amssymb}
\usepackage{stmaryrd}
\usepackage{graphicx}
\usepackage{fancyvrb}
\usepackage{amsthm}
\usepackage{mathtools}
\usepackage{bm}
\usepackage{array}
\usepackage{enumitem}
\usepackage{proof}

\theoremstyle{definition} % amsthm only
\newtheorem{definition}{Definition}

\graphicspath{{figures/}}

\DeclareMathOperator{\lk}{\ell}

\newcommand{\soft}[1]{\mathrel{\tilde{#1}}}
\newcommand{\softv}[1]{\tilde{#1}}
\newcommand{\dsoft}[1]{\mathrel{\tilde{#1}}}

\usepackage{accents}

\newcommand{\norm}[1]{\left\|{{#1}}\right\|}
\DeclareMathOperator{\cond}{cond}

\newcommand{\cM}{{\mathcal{M}}}

\usepackage[colorinlistoftodos,prependcaption,textsize=tiny]{todonotes}
\usepackage{xargs} % Use more than one optional parameter in a new commands

\newcommandx{\unsure}[2][1=]{\todo[linecolor=red,backgroundcolor=red!25,bordercolor=red,#1]{#2}}
\newcommandx{\change}[2][1=]{\todo[linecolor=blue,backgroundcolor=blue!25,bordercolor=blue,#1]{#2}}
\newcommandx{\info}[2][1=]{\todo[linecolor=OliveGreen,backgroundcolor=OliveGreen!25,bordercolor=OliveGreen,#1]{#2}}
\newcommandx{\improvement}[2][1=]{\todo[linecolor=Plum,backgroundcolor=Plum!25,bordercolor=Plum,#1]{#2}}
\newcommandx{\wtp}[2][1=]{\todo[linecolor=orange,backgroundcolor=orange!25,bordercolor=orange,#1]{#2}}
\newcommandx{\thiswillnotshow}[2][1=]{\todo[disable,#1]{#2}}

\makeatletter
\newenvironment{exprogram}[1][htb]{%
    \renewcommand{\ALG@name}{Example Program}% Update algorithm name
   \begin{algorithm}[#1]%
  }{\end{algorithm}}
\makeatother

\makeatletter
\newcommand{\ALOOP}[1]{\ALC@it\algorithmicloop\ #1%
  \begin{ALC@loop}}
\newcommand{\ENDALOOP}{\end{ALC@loop}\ALC@it\algorithmicendloop}
\makeatother

\renewcommand{\algorithmicloop}{\textbf{subroutine}}

\icmltitlerunning{Soft Constraints for Inference with Declarative Knowledge}
\begin{document}

\twocolumn[
\icmltitle{Soft Constraints for Inference with Declarative Knowledge}



\icmlsetsymbol{equal}{*}

\begin{icmlauthorlist}
\icmlauthor{Zenna Tavares}{mit}
\icmlauthor{Javier Burroni}{amhert}
\icmlauthor{Edgar Minaysan}{princeton}
\icmlauthor{Armando Solar Lezama}{mit}
\icmlauthor{Rajesh Rangananth}{nyu}
\end{icmlauthorlist}

\icmlaffiliation{mit}{MIT, USA}
\icmlaffiliation{nyu}{NYU, USA}
\icmlaffiliation{amhert}{UMass Amherst, USA}
\icmlaffiliation{princeton}{Princeton University, USA}

\icmlcorrespondingauthor{Zenna Tavares}{zenna@mit.edu}
% \icmlcorrespondingauthor{Eee Pppp}{ep@eden.co.uk}

\icmlkeywords{Probabilistic Inference, Markov Chain Monte Carlo, Replica Exchange, Probabilistic Programming}

\vskip 0.3in
]



\printAffiliationsAndNotice{\icmlEqualContribution} % otherwise use the standard text.

\begin{abstract}
We develop a likelihood free inference procedure for conditioning a probabilistic model on a predicate.
A predicate is a Boolean valued function which expresses a yes/no question about a domain.
% when conditioned on, they enforce propositions such as ``rigid bodies do not intersect``, or observations that coarse e.g., that a person is tall, when height in inches is explicit in the model; or even complex logical combinations.
% Conditioning on a predicates remains severely limited due to intractable likelihoods.
Our contribution, which we call predicate exchange, 
constructs a softened predicate which takes value in the unit interval [0, 1] as opposed to a simply true or false. Intuitively, 1 corresponds to true, and a high value (such as 0.999) corresponds to ``nearly true'' as determined by a distance metric.
We define Boolean algebra for soft predicates,  such that they can be negated, conjoined and disjoined arbitrarily.
A softened predicate can serve as a tractable proxy to a likelihood function for approximate posterior inference.
However, to target exact inference, we temper the relaxation by a temperature parameter, and add a accept/reject phase use to replica exchange Markov Chain Mont Carlo, which exchanges states between a sequence of models conditioned on predicates at varying temperatures.
We describe a lightweight implementation of predicate exchange that it provides a language independent layer that can be implemented on top of existingn modeling formalisms.
\end{abstract}

% !TEX root = icmlsoft.tex

\section{Introduction}

Conditioning in Bayesian inference incorporates observed data into a model.
In a broader sense, conditioning revises a model such that a yes/no question (a predicate) is resolved to a true proposition (a fact).
For instance, the question of whether a variable is equal to a particular value, changes from a predicate of uncertain truth, to a fact, once it is observed.
In principle, a predicate can be used to declare any fact about a domain, not only the observation of data.
In practice, sampling from models conditioned on most predicates presents severe challenges to existing inference procedures.

Predicates can be used to update a model to adhere to known facts about a domain, without the burden of specifying how to revise the model.
For example, in inverse graphics \cite{marschner1998inverse,kulkarni2015deep} (inferring three dimensional geometry from observed images), the proposition ``rigid bodies do not intersect'' is a predicate on latent configurations of geometry.
To manually revise a model to constructively adhere to this fact is ranges between inconvenient and infeasible.
Instead, we would ideally simply condition on it being true, concentrating probability mass on physically plausible geometric configurations, ultimately to yield more accurate posterior inferences in the inverse graphics problem.

Predicates can also express observations that are more abstract than variables in a model.
In diabetes research for example, probabilistic models have been used to relate physiological factors to glucose levels over time \citep{levine2017offline,murata2004probabilistic}.
Rather than concrete, numerical glucose measurements, a medical practitioner may observe (or be told) that a patient suffers from recurrent hypoglycemia, i.e., that their glucose levels periodically fall below a critical value.
Even if the occurrence of hypoglycemia does not appear as an explicit variable in the model, it could be constructed as a predicate on glucose levels, and conditioned on to infer the posterior distribution over latent physiological factors.

Several effective sampling  \citep{andrieu2003introduction} and variational  \citep{jordan1999introduction, ranganath2014black} approaches to inference require only a black-box likelihood function, i.e., one evaluable on arbitrary input.
The likelihood function quantifies the extent to which values of latent variables are consistent with observations. 
However, most models conditioned on most predicates have likelihood functions that are intractable to compute or unknown.
For example, conditioning random variables that are deterministic transformations of other random variables (e.g., the presence of hypoglycemia in the example above, or the mean of a collection of variables) often results in likelihoods that are normalized by intractable integrals.	
% For example, given three variables, it is possible to condition on the sum
% of the variables being positive; this sum is never an explicit random variable.
In other cases, the likelihood function is implicit to a generative process, rather than explicitly specified, and hence unavailable even when the condition is a conventional observation.

In this paper we present predicate exchange:
a likelihood-free method to sample from distributions conditioned on predicates from a broad class.
It is composed of two parts:
\begin{enumerate}
\item \textbf{Predicate Relaxation} transforms a predicate such that it returns a value in a soft Boolean algebra: the unit interval $[0, 1]$ with continuous logical connectives $\soft{\land}$. $\soft{\lor}$ and $\neg$.
\item  \textbf{Replica Exchange} simulates several Markov chains of a model at different temperatures.  Temperature is a parameter of predicate relaxation which controls the amount of approximation it introduces.  We adapt standard replica exchange to draw samples that are asymptotically exact from the unrelaxed model. 
\end{enumerate}

% WTP? Show that soft Boolean is useful for (i) tractability of inference show that rather th	an 
By returning a value in $[0, 1]$ instead of $\{0, 1\}$, a soft predicate quantifies the extent to which values of latent variables are consistent with the predicate.
This allows it to serve a role similar to a likelihood function, and opens up the use of likelihood-based inference procedures.
Orthogonally, we embed $]0, 1]$ in a Boolean algebra to support the expression of domain knowledge of composite Boolean structure.
Continuing the previous example, we may know that a person does \emph{not} have hypoglycemia, or that they have hypoglycemia \emph{or} hyperglycemia, or \emph{neither}.
% Few likelihood-based likelihood-free approaches to inference offer effective means to condition non-trivial models on compound predicates.

% This degree to which a is is determined by a notion of distance.
% In contrast to most distance based inference methods (notably Approximate Bayesian Computation \cite{beaumont2002approximate}), we develop a form of replica exchange Markov Chain Monte Carlo \cite{earl2005parallel} to target inference that is exact in convergence of the chain.
% Predicate relaxation is modulated by temperature such that at zero temperature the relaxed predicate mirrors its hard counter-part, while at maximal temperatures, it is virtually always satisfied.
% Predicate Exchange simulates several Markov chains in parallel at different temperatures.

Predicate exchange is motivated by probabilistic programming languages, which have vastly expanded the class of probabilistic models that can be expressed,, but still heavily restrict the kinds of predicates that can be conditioned on.
Rather than introduce a new language or modeling formalism, we mirror   \cite{wingate2011lightweight} and provide a light-weight implementation that performs inference by modulating the execution of a stochastic simulation based model.
This means predicate exchange is easily incorporated into most frameworks. 

Our approach comes with certain limitations.
Equality conditions on continuous variables indicate sets of zero measure.
This is problematic because the probability of proposing a satisfying state in a Markov chain becomes zero.
In these cases predicate exchange must sample at a minimum temperature strictly greater than zero, which is approximate.
Another limitation occurs if a predicate has branches (e.g., if-then-else statements) which depend on uncertainty in the model.
% Since all possible paths through a program are visited, our measure of is necessarily an estimate.

In summary we address the problem of conditioning probabilistic models on predicates as a means to express declarative knowledge.
In detail, we:

\begin{enumerate}
	\item Formalize simulation based probabilistic models in measure theoretic probability, and conditioning as the imposition of constraints expressed as predicates (Section \ref{simmodels}).
	\item Motivate predicate relaxation (Section \ref{predexchange}), and provide a complete soft Boolean algebra. 
	\item Provide a light-weight implementation of predicate exchange (Section \ref{implement}) through nonstandard execution of a simulation based model.
	\item Evaluate our approach on examples, including a case study in glycemic forecasting.
\end{enumerate}

% WTP: Contribution: inference algorithm that supports conditioning on a wider class of propositions
% In this paper we present an algorithm that draws samples from generative models that have been conditioned on predicates belonging to a more general class than observation of data.
% Predicates, when used as black boxes, provide only sparse information -- the constraint is satisfied or it is not -- and the subset of satisfying constraints is typically vanishingly small.
% Our objective is to support conditioning on predicates on spaces for which a natural metric can be defined.
% A metric provides more information a measure of the degree of satisfaction, and allows us.

% WTP: Paper summary
% In summary we address the problem of conditioning on declarative knowledge.
% In more detail:
% \begin{itemize}
% \item We formalize simulation models in measure-theoretic probability as random variables defined on a shared probability space (section X), and define conditioning as a concentration of measure.
% \item We describe our approach to inference, which softens the hard constraints to admit tractable inference in a broader set of scenarios.
% \item  We demonstrate our approach on a number of examples, with experiments on toy data and experiments on medical models by enriching them with declarative knowledge to learn from limited data.
% \end{itemize}
\section{Related Work}

% Likelihood free inference
Demand for likelihood-free inference emerged in genetics ecology.
Tavar{\'e} et al. \yrcite{tavare1997inferring} compared summary statistics of the output of a simulation with that of observed data, and rejected mismatches.
%  to perform one of the earliest forms of likelihood-free inference.
Weiss et al. \yrcite{weiss1998inference} expanded on this with a tolerance term, so that simulations yielding data sufficiently close to the targets were accepted.
% Such posterior samples are therefore approximate.
Approximate Bayesian Computation (ABC) has come to refer to broad class of methods \cite{beaumont2002approximate,sisson2007sequential} in this general regime.
Marjoram et al. \yrcite{marjoram2003markov} simulated Markov Chains according to the prior, but introduced the accept/reject stage to yield approximate posterior samples.
A small tolerance leads to a high rejection rate, whereas a large tolerance results in an unacceptable approximation error.
Among several solutions are dynamically decreasing the tolerance \cite{toni2008approximate}, importance reweighting samples based on distance \cite{wegmann2009efficient}, adapting the tolerance based on distance \cite{del2012adaptive,lenormand2013adaptive}, as well as annealing the tolerance as a temperature parameter \cite{albert2015simulated}.

Predicate exchange targets simulation models and uses distance metrics, but targets exact inference without summary statistics.
A recent approach \cite{graham2017asymptotically}  with similar objectives develops a Hamiltonian Monte Carlo variant, using a quasi-Newton method during leap-frog integration to exactly solve the observation constraint.
This is limited to differentiable models conditioned with equality.

% \cite{Pseudo-Marginal Hamiltonian Monte Carlo}HMC methods cannot be implemented in scenarios where the likelihood function is intractable. However, we have shown here that if we have access to a non-negative unbiased likelihood estimator.
% parameterized by normal random variables then it is possible to derive an algorithm which mimicsthe HMC algorithm having access to the exact likelihood. The resulting pseudo-marginal HMCalgorithm replaces the original intractable gradient of the log-likelihood by the gradient of thelog-likelihood estimator while preserving the target distribution as invariant distributi

% PPLS
Probabilistic logics such as ProbLog \cite{richardson2006markov} and Markov logic networks \cite{de2007problog} allow extend first order logic to declare both models and conditions.
More recent probabilistic programming systems~\citep{milch20071, wood2014new,mansinghka2014venture,goodman2008church,carpenter2017stan} have focused on stochastic simulation, and automatically automatically derive the likelihood function for a rich class of models.

% ?
% Our approach is related to smooth interpretation of programs \citep{chaudhuri2010smooth}

Several continuous \cite{levin2000continuous} and fuzzy \cite{klir1995fuzzy} logics apply model-theoretic tools to metric structures.
Continuous logics replace the Boolean structure $\{T, F\}$, quantifiers $\forall x$ and $\exists x$, and logical connectives with continuous counter-parts.
Predicate relies uses a continuous logic only make inference more tractable. Semantically, our approach remains within measure theoretic foundations, which relies on hard predicates to condition.
% Probabilistic Similarity Logic \cite{brocheler2012probabilistic,kimmig2012short} uses continuous logics 
% Talk about Soft logic

% !TEX root = nips_2018.tex

\section{Simulation Models}\label{simmodels}

% The density function of the resulting model, if it exists, is not explicitly represented, but rather implicitly defined.
% For this reason simulator models are often called implicit or generative models.

% Simulator-based models are useful because they interface easily with models typicallyencountered in the natural sciences. In particular, hypotheses of how the observed datayowere generated can be implemented without making excessive compromises in order tohave an analytically tractable model

Probabilistic simulation based models specify the step-by-step causal mechanisms of a domain, and use probability distributions for any uncertain parameters.
A simulation model can be stochastically executed, using a random number generator to sample from primitive random variables in the model.
Inference means to simulate the model while imposing constraints on variables in the model.
This is difficult, since simulation based models lack an explicit likelihood function, which is necessary for most inference procedures.

Conditioning on predicates requires a measure-theoretic foundation, in which a simulation model is a random variable:

\begin{figure}
	\centering
	\begin{minipage}[t]{0.5\linewidth}
		\centering
		\includegraphics[width=0.8\linewidth]{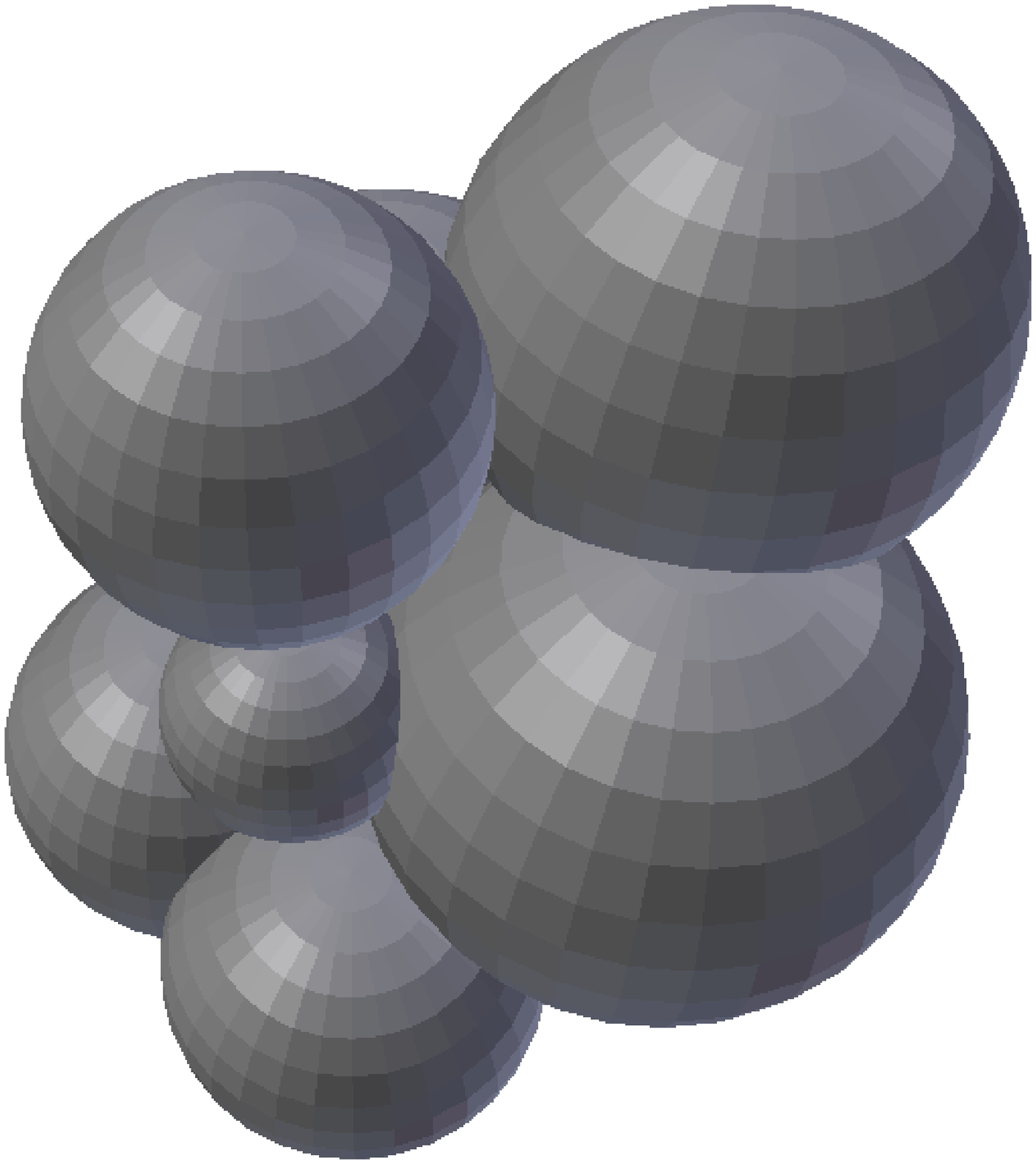}
	\end{minipage}%
	\begin{minipage}[t]{0.5\linewidth}
		\centering
		\includegraphics[width=0.8\linewidth]{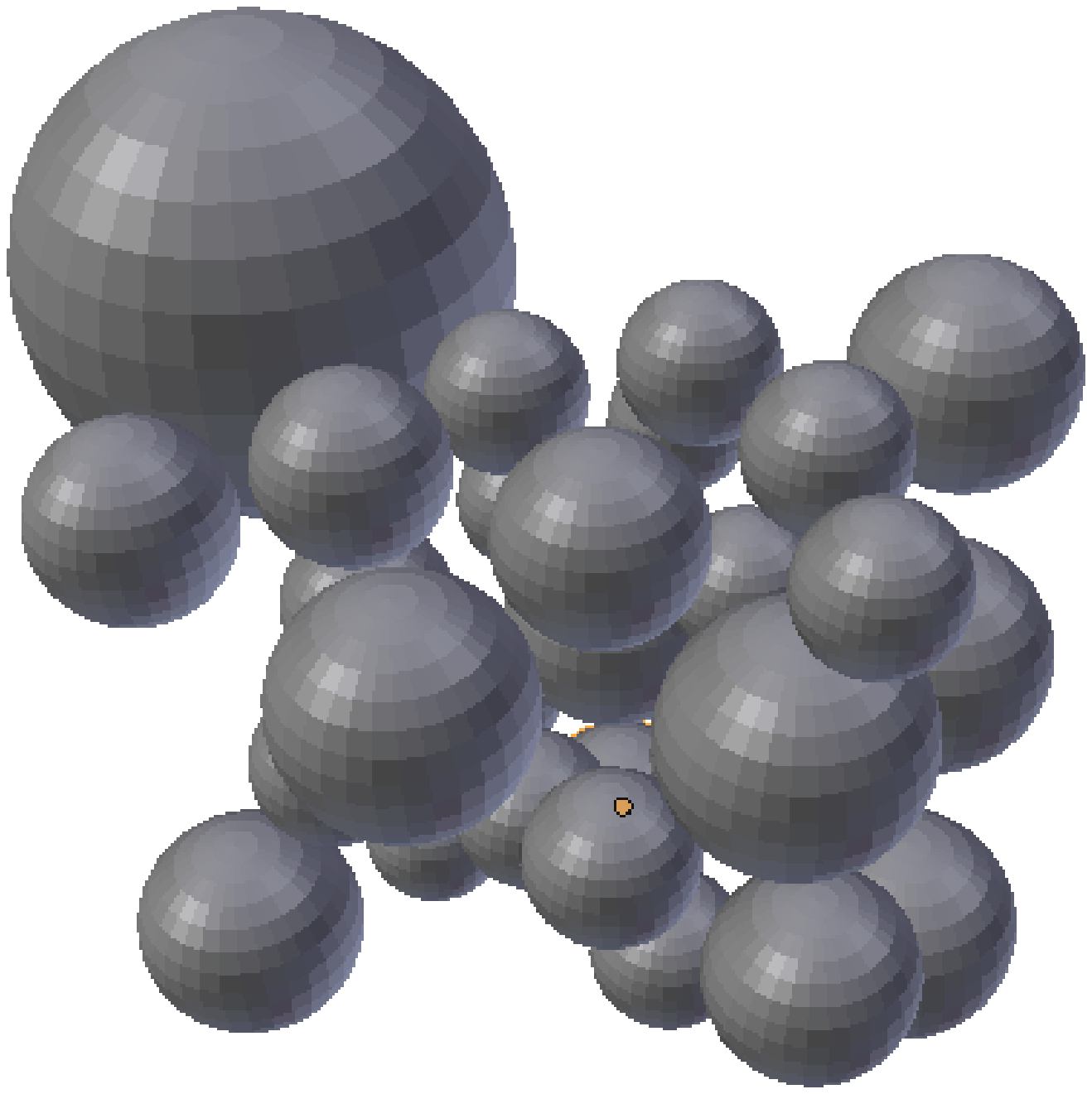}
	\end{minipage}%
	\caption{Sample from geometric prior (left), whereas (right) is conditioned on no-intersection constraint}
	\label{fig:nointersect}
\end{figure}

\paragraph{Random Variables.} Probability models lie on top of probability
spaces. A probability space is a measure space $(\Omega, {\cal H}, {\cal P})$,
where ${\cal H}$ is a sigma algebra and ${\cal P}(\Omega) = 1$ \citep{ccinlar2011probability}. Random variables are
functions from the space $\Omega$ to a realization space ${\cal X}$. As a concrete
example the space $\Omega$ can be thought of as a hypercube, with ${\cal P}$ being
uniform over that hypercube. To build a normal random variable, we need a function
that maps from $\Omega \to \mathbb{R}$. If the underlying probability space is uniform, then
this function is the inverse cumulative distribution function of the normal.

A \emph{model} $\cM$ is a collection of random variables along with a probability space.
% To fully specify a model, we need both the measure space and the
% collection of random variables on which it acts.

\paragraph{Conditioning}

Conditioning a model creates a new model.
As an example consider a model $\cM$ with two
random variables $X_1$ and $X_2$ that both take real values. Conditioning
$\cM$ on $X_1 = 1$, defines a new model $\cM_{|A}$ based on limiting the measure space
$\Omega$ to the set $A = \{ \omega : X_1(\omega) = 1\}$.
The new model is defined on a new probability space
\begin{align}
	(\Omega \cap A, \{A \cap B, B \in {\cal H} \}, {\cal P} / {\cal P}(A))
\end{align}
with the same random variables $X_1$ and $X_2$.
Sampling from $\cM_{|A}$ produces
samples only where $X_1 = 1$

More generally, conditioning on any predicate $Y(\omega) = \lk(X_1(\omega), \dots, X_n(\omega))$
defines a new model defined exactly as above, where $A = \{\omega : \lk(X_1(\omega), \dots, X_n(\omega)) = 1\}$.
% This predicate may include a comparisonsuch as $X_i < X_j$, restrict deterministic function of variables in in the model such as $\exp(X_i) = 2$, or be a Boolean combination such as $(X_i < X_j) \lor \neg(\exp(X_i) = 2)$.
Sampling from $\cM_{|A}$ generates $(x_1, ..., x_n)$ where $\lk$ is true.

The general construction of new models might require conditioning
on sets of measure zero. This process can be made rigorous
via disintegration \citep{chang1997conditioning}. Disintegration can
be thought of as the reversal of building joint distributions through
product measure constructions.

\section{Predicate Exchange}
To condition a model $\cM$ on a predicate $Y$ we develop \emph{predicate exchange}, a likelihood-free inference procedure.  It is composed of two parts:
\begin{enumerate}
\item \textbf{Predicate Relaxation} constructs a soft predicate $\softv{Y}$ from $Y$. $\softv{Y}$ takes values   in a soft Boolean algebra: the unit interval $[0, 1]$ with continuous logical connectives $\soft{\land}$. $\soft{\lor}$ and $\neg$. 
$\softv{Y}$ is 1 iff $Y$ is 1, but otherwise takes nonzero values denoting the degree to which $Y$ is satisfied.
% The relaxation is parameterized by a temperature.
% Predicate relaxation allows us to apply likelihood based MCMC inference procedures.
\item  \textbf{Replica Exchange} is a Markov Chain Monte Carlo procedure that exploits temperature. The strength by which $\softv{Y}$ relaxes $Y$ is modulated by a temperature parameter $\alpha$, which trades off between accuracy and ease of inference.  By simulating several replicas of $\softv{Y}$ at different temperatures, replica exchange is able to draw exact samples. 
\end{enumerate}

% \subsection{Soft Conditioning}
% Conditioning is an operation that restricts a measure space by a set.

% Let $(\Omega, \mathcal{H}, \mathcal{P})$ be a measure space.
% Soft conditioning on a function $Y: \Omega \to [0, 1]$ constructs a new measure $\mathcal{P}'$ defined as:
% \begin{equation}
% \mathcal{P}'(A) = \frac{\mathbb{E}(Y \mid A)}{\mathbb{E}(Y)}
% \end{equation}

\subsection{Predicate Relaxation}\label{predexchange}

A soft predicate $\softv{Y}$ approximates $Y$ in the sense that when viewed as a likelihood function on model parameters, $\softv{Y}$ has a broader support, assigning nonzero weights to parameter values which have zero weight under $Y$.
There are three desiderata which govern this approximation.
First, $\softv{Y}$ should have a temperature parameter $\alpha$ that controls the fidelity of the approximation. In particular, $\softv{Y}$ should converge to $Y$ as $\alpha \to 0$, and to a flat surface as $\alpha \to \infty$. Second, the fidelity of the approximation should vary monotonically with temperature. Third, $\softv{Y}$ should be consistent with $Y$ on 1. That is $Y(\omega) = 1$ iff $\softv{Y}(\omega) = 1$ at all temperatures.

% The family of approximations of the predicate $Y$ is parameterized through a temperature $\alpha$ that controls the smoothness of the approximation. In particular, $\softv{Y}$ with $\alpha \to 0$ converges to $Y$ itself while increasing values of $\alpha$ yield smoother approximations eventually giving a flat surface when $\alpha \to \infty$. Moreover, the set of conditional samples $C(Y) = \{ \omega \in \Omega \text{ } | \text{ } Y(\omega) = 1 \}$ are assigned a value of $1$ in $\softv{Y}$ for all temperatures. 

% To construct a $\softv{Y}$ with such properties, we let $Y$ be the following predicate $a=b$ for standard Gaussians $a, b \sim \mathcal{N}(0, 1)$. We choose a distance
% $\rho(a, b)$ to indicate how close our sample is to $C(Y)$. To meet the desiderata
% of having $C(Y)$ be $1$ over the constraint set and to ensure the constraint becomes
% smoother with large $\alpha$, we use a function $k : [0, \infty] \to [0, 1]$ parameterized by $\alpha$ to wrap the distance $\rho(a, b)$. A simple choice for such a function is $k(d; \alpha) = e^{-d / \alpha}$ that provides the desired properties to $\softv{Y}$ (see Appendix). The formal definition of $\softv{Y}$ is as follows.

\begin{definition}
A function $\softv{Y} : \Omega \to [0, 1]$ parameterized by $\alpha \in [0, \infty)$ is a relaxation of $Y: \Omega \to \{0, 1\}$ if:
\begin{enumerate}[label=(\roman*)]
	\label{def:temp}
	\item For all $\omega \in \Omega$, $\lim_{\alpha \to 0}\softv{Y}(\omega; \alpha) = Y(\omega)$.
	\item For all $\omega \in \Omega$, $\lim_{\alpha \to \infty}\softv{Y}(\omega; \alpha) = 1$.

    \item For all $\alpha$, $\softv{Y}(\omega; \alpha) = 1$ iff $Y(\omega) = 1$.
    \item The entropy $H(\softv{Y}(\omega; \alpha))$ (which characterizes the fidelity of the approximation ) is an increasing function of $\alpha$.\footnote
    {By compactness, it is integrable for all $\alpha$, when $\Omega$ has finite dimension}
\end{enumerate}
\end{definition}

\paragraph{Graded Satisfiability}
$\soft{\lk}_{\inf}(m)$ represents the degree to which a model realization $m$ satisfies a predicate.
Let $k_\alpha$ be a kernel (described below), $\rho$ a distance metric, and $A = \{(x_1, \dots, x_n) \mid \lk(x_1, \dots, x_n) = 1\}$ the satisfying set.
$\soft{\lk}_{\inf}(m)$ is then:
\begin{equation}
\soft{\lk}_{\inf}(m) = k_\alpha(\rho(m, A))
\end{equation}
where $\rho(x, A) = \inf \left\{\rho(x, a) \mid a \in A\right\}$.
% As shown in the Section \ref{implement}, $\soft{\lk}_{\inf}$ can be difficult to compute.

\paragraph{Distance} A relaxation kernel $k_\alpha$ bounds distances from $\rho$ to the unit interval, and is paramterized by temperature $\alpha$.
% For example if $x$ and $y$ are real values, then $x \soft{=} y$ is defined as $k_\alpha(\rho(x, y))$ where $\rho$ is a distance function and $k_\alpha$ is a relaxation kernel paramterized by temperature $\alpha$.
% A relaxation kernel maps distances to values in $[0, 1]$, and ensures that $\softv{Y}$ adheres to the outlined criteria.
We restrict our attention to the squared exponential kernel:
\begin{equation}
k_{\alpha}(r) = \exp\left(-\frac{r^2}{\alpha}\right)
\end{equation}

$\rho$ is parameterized by the type of input.
For canonical spaces such as $\mathbb{R}$ and $\mathbb{N}$ we default to the Euclidean distance. 
$x \soft{=} y$ is then defined as $\exp(\norm{x - y}/\alpha)$.
For composite elements $x, y \in \mathbb{T}_1 \times \cdots \times \mathbb{T}_n$ of product type, by default $\rho$ takes a mean $\rho(x, y) = (1/n)\sum^n_{i=1}\rho(x_i, y_i)$.

  % \begin{center}
  % \begin{tabular}{ c |  c | c }
  %   \hline		
  %   $x \soft{=} y$ & $x \soft{>} y$ & $x \soft{<} y$  \\
  %   $k_\alpha(\rho(x, y))$ & $k_\alpha(\rho(x, [y, \infty]))$ & $k_\alpha(\rho(x, [-\infty, y]))$ \\
  %   \hline  
  % \end{tabular}
  % \end{center}

\paragraph{Composition} We construct $\softv{Y}$ from $Y$ compositionally, by substituting primitive predicates (equality, inequalities and logical operators) with soft counterparts.
For instance the predicate $(x > y) \lor \neg(x^2 = 2)$ is transformed into $(x \soft{>} y) \soft{\lor} \softv{\neg}(x^2 \soft{=} 2)$.
In general, we use $\soft{p}$ to denote a relaxation of a predicate $p$.
\begin{figure}[H]\label{softpreds}
  \begin{align*}
x \soft{=} y &= k_\alpha(\rho(x, y))\\
x \soft{>} y &= k_\alpha(\rho(x, [y, \infty]))\\
% x \soft{<} y &= k_\alpha(\rho(x, [-\infty, y]))\\
x \soft{<} y &= k_\alpha(\rho(y, [-\infty, x]))\\
a \soft{\land} b &= \max(a, b)\\
a \soft{\lor} b &= \min(a, b)
\end{align*}
\caption{Soft Primitive Predicates}
\end{figure}

A soft inequality such as $x \soft{>} y$ is function of the amount by which $x$ must be increased (or $y$ decreased) until $x > y$ is true.
This is the distance between $x$ and the interval $[y, \infty]$, where the distance between a point and any interval $[a, b]$ is the smallest distance between $x$ and any element in $[a, b]$, and therefore 0 if $x \in [a, b]$:
\begin{equation}
\rho(x, [a, b]) =
\begin{cases}
  a - b & \text{ if } x < a\\
  x - b & \text{ if } x < b\\
  0              & \text{otherwise}
\end{cases}
\end{equation}

\begin{figure}
\includegraphics[width=\linewidth]{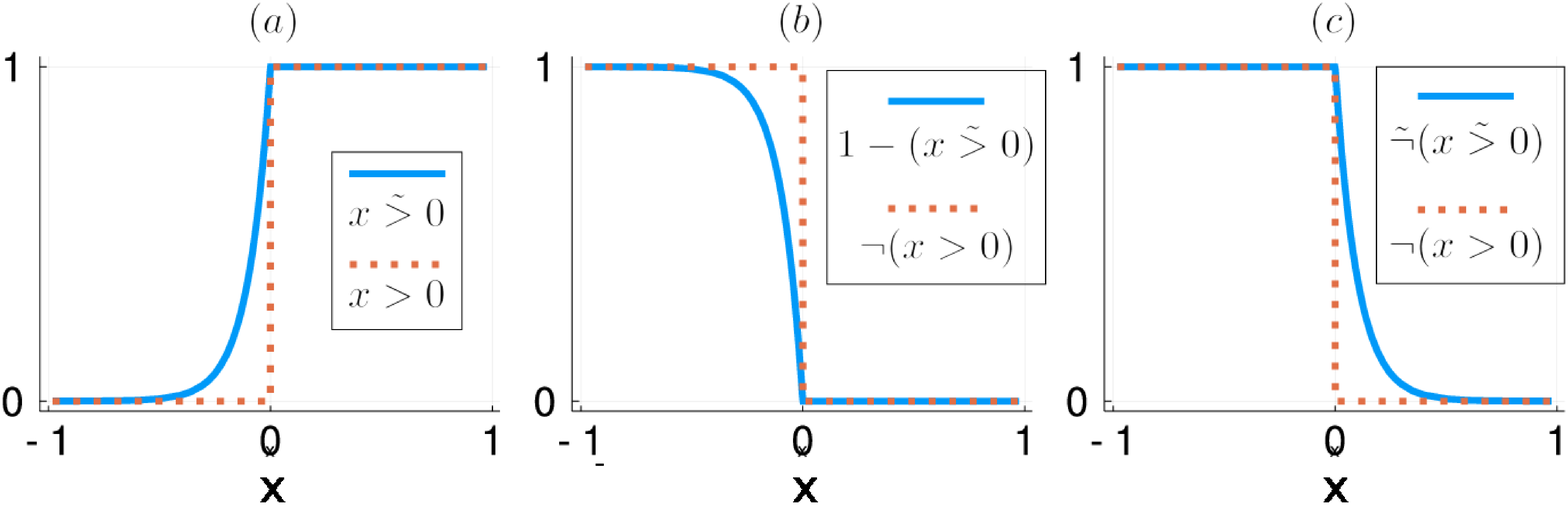}
\caption{Soft predicates as function of $x$.  In all figures the blue line denotes the soft predicate, while the red line denotes the predicate to approximate.}\label{negationimg}
\end{figure}

Soft negation introduces complications.
To illustrate, Figure \ref{negationimg} (a) shows $x \soft{>} 0$ as a function of $x$.
In continuous logics \cite{kimmig2012short}, the negation of $a \in [0, 1]$ is $1 - a$.
However, as shown in Figure \ref{negationimg} (b), this violates criteria (iii) of predicate relaxation; there are values which satisfy the hard predicate $\neg(x > 0)$ which do take a value of 1 in $1 - (x \soft{>} 0)$.

The problem of negation arises because $\softv{Y}$ is consistent with $Y$ at 1 but not at 0.
In other words, $\softv{Y}$ is a one-sided approximation.
To overcome this challenge, soft primitives yield a pair $(a_0, a_1)$ where $a_0, a_0 \in [0, 1]$.
$a_1$ preserves consistency with $Y$ on $1$, just as before, while $a_0$ preserves consistency with $\neg Y$ on $1$.
For example if $x \dsoft{>} 0 = (a_0, a_1)$, then as a function of $x$, $a_0$ and $a_1$ correspond to Figure \ref{negationimg} (a) and (c) respectively.

A complete two-sided soft logic is shown in Figure \ref{softw}.
Although a two-sided predicate has two components, for the sake of conditioning we are still concerned only with the true side $a_1$ in the pair $(a_0, a_1)$.
Soft negation simply swaps the elements of $(a_0, a_1)$ to yield $(a_1, a_0)$.

% \begin{definition}
% The function $\softv{Y} : \Omega \to [0, 1]^2$ parameterized by $\alpha \in [0, \infty)$ is a two-sided relaxation of a $Y: \Omega \to \{0, 1\}$ if:
% \begin{enumerate}[label=(\roman*)]
% 	\label{def:temp}
% 	\item For all $\omega \in \Omega$, $\lim_{\alpha \to 0}\softv{Y}(\omega; \alpha) = (\neg Y(\omega), Y(\omega))$.
% 	\item For all $\omega \in \Omega$, $\lim_{\alpha \to \infty}\softv{Y}(\omega; \alpha) = (0, 1)$.

%     \item For all $\alpha$, $\softv{Y}(\omega; \alpha) = 1$ iff $Y(\omega) = 1$.
%     \item The entropy $H(\softv{Y}(\omega; \alpha))$ (which characterizes the fidelity of the approximation ) is an increasing function of $\alpha$.\footnote
%     {By compactness, it is integrable for all $\alpha$, when $\Omega$ has finite dimension}
% \end{enumerate}
% \end{definition}

\begin{figure}
\begin{align*}
x \dsoft{=} y &= (\text{if } x = y  \text{ then } \exp(1/\alpha) \text{ else } 1, k_\alpha(\rho(x, y)))\\
x \dsoft{>} y &= (k_\alpha(\rho(x, [-\infty, y])), k_\alpha(\rho(x, [y, \infty])))\\
x \dsoft{<} y &= (k_\alpha(\rho(y, [x, \infty])), k_\alpha(\rho(y, [-\infty, x])))\\
(a_0, a_1) \dsoft{\land} (b_0, b_1) &= (a_0 \soft{\land} b_0, a_1 \soft{\land} b_1)\\
(a_0, a_1) \dsoft{\lor} (b_0, b_1) &= (a_0 \soft{\lor} b_0, a_1 \soft{\lor} b_1)\\
\softv{\neg}(a_0, a_1) &= (a_1, a_0)
\end{align*}
\caption{Two sided soft primitive predicates}
\label{softw}
\end{figure}

% \begin{center}
% \begin{tabular}{ l | c | r }
%   % \hline		
%   $(a_0, a_1) \soft{\land} (b_0, b_1)$ & $(a_0, a_1) \soft{\lor} (b_0, b_1)$ & $\neg(a_0, a_1)$ \\
%   $(a_0, a_1) \soft{\land} (b_0, b_1)$ & $(a_0, a_1) \soft{\lor} (b_0, b_1)$ & $\neg(a_0, a_1)$ \\
%   % \hline  
% \end{tabular}
% \end{center}

\paragraph{Unsatisfiability}Predicate exchange is unable to determine if a predicate is unsatisfiable (e.g.  $(x > 1) \land (x < -1)$), and defers to the user to ensure this is the case.

\subsection{Approximate Markov Chain Monte Carlo}
A soft predicate can serve as an approximate likelihood, and as a result is amenable to likelihood based inference methods such as Markov Chain Monte Carlo.
MCMC algorithms require a function $f$ that is proportional to the the target density.
In Bayesian inference this is the posterior, dictated by Bayes' theorem as the product of the likelihood and the prior.
Approximate inference using soft predicates takes a similar form.

Let $\cM = (X_1, \dots, X_n)$ be a model, $Y$ be a predicate that conditions $\cM$, and  $\softv{Y}(\omega) = \softv{\lk}(X_1(\omega), ..., X_n(\omega))$ be a relaxation of $Y$.
% % Random variables in $\cM$ may be exogenous or endogenous in the sense that
% % given values for exogenous variables, the values of endogenous are determined.
% % For example $X = \mathcal{N}(0,1)$ is exogenous while $Y = X^2$ is endogenous.
% % Inference need only take place on exogenous random variables.
% Both $Y$ and $\softv{Y}$ are random variables, and hence map from the sample space $\Omega$;
% for convenience we define a soft predicate $\softv{\lk}$ as a function of its parameters (other random variables in the model) rather than the sample space.
% That is, let $\softv{Y}(\omega) = \softv{\lk}(X_1(\omega), ..., X_n(\omega))$, where $X_i \in \cM$.
Assuming a prior density $p$, the approximate posterior $f$ is the product:
\begin{equation}
f(m) = p(m) \cdot \softv{\lk}(m)
\end{equation}
$\softv{\lk}$ down weights parameter values which violate $Y$ by the degree to which they violate it. 
This is modulated by the temperature $\alpha$ used in the  relaxation kernels which constitute $\softv{\lk}$.
At maximum temperature $\softv{\lk}$ has no effect, and the approximate posterior $f$ is equal to the prior $p$.
At zero temperature, $f$ recovers the true posterior since parameter values which violate the condition are given zero weight.

\begin{figure}
\includegraphics[width=\linewidth]{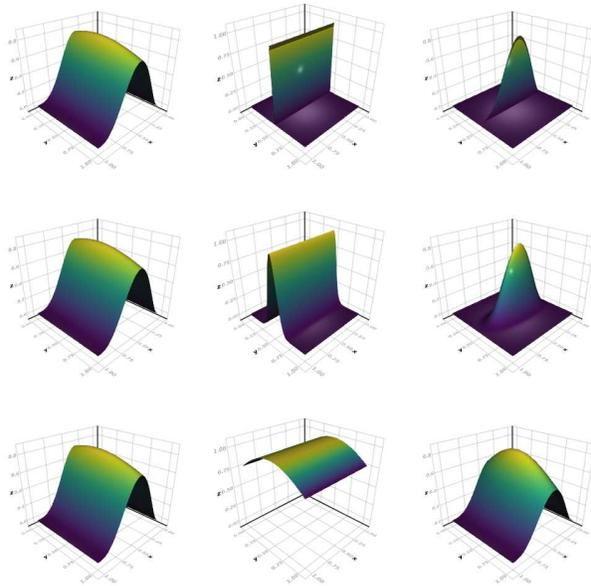}
\caption{Approximate Posterior at varying temperatures.  Temperature decreases from top row to bottom.  Along each row: (left) is the prior term $p$, (center) is the soft likelihood term $\softv{\lk}$, and (right) is the approximate posterior $f$}
\label{temppost}
\end{figure}

For illustration, let $\cM = (\mu, X)$ be a model where $\mu = \beta(3, 4), X = \mathcal{N}(\mu, 1)$  conditioned on $X = 0.5$.
The approximate posterior is shown at different temperatures in Figure \ref{temppost} and defined as:
\begin{equation}\label{approxposterior}
f_\alpha(\mu, x) = \beta_{0,1}(\mu) \cdot \mathcal{N}_{\mu,1}(x) \cdot k_\alpha(\rho(x, 0.5)) 
\end{equation}

The temperature parameter trades off between tractability of inference and the fidelity of the approximation.
Too high and $\softv{Y}$ will diverge too greatly from $Y$. Too low and convergence will be slow.
% Several temperatre methods for controlling temperature exist such as simulated annealing \citep{kirkpatrick1983optimization} and Adiabatic Monte Carlo \citep{betancourt2014adiabatic}.
% We augment replica exchange , which combines information from several chains to sample more effectively, which an accept/reject phase, to sample from the unrelaxed model.

\subsection{Replica Exchange}\label{replicaexchange}
Replica exchange simulates \citep{swendsen1986replica} $M$ replicas at different temperatures, and uses a Metropolis-Hastings update to  periodically swap the temperatures of chains.
If $f_{\alpha_i}$ is an approximate posterior function at temperature $\alpha_i$, two independent parallel chains simulating targets $f_{\alpha_1}(x)$, $f_{\alpha_2}(y)$ they follow a joint target $f_{\alpha_1, \alpha_2}(x,y) = f_{\alpha_1}(x)f_{\alpha_2}(y)$.
Replica exchange swaps states between the chains while preserving the joint target.
Swapping states is equivalent to swapping predicates, which motivates the name predicate exchange.
Concretely, replica exchange proposes a swap from $(x, y)$ to $(y, x)$, and accepts it with probability $\min(1, A)$, where:
\begin{equation}
A =  \frac{f_{\alpha_1, \alpha_2}(y,x)}{f_{\alpha_1, \alpha_2}(x,y)} = \frac{f_{\alpha_1}(y)f_{\alpha_2}(x)}{f_{\alpha_1}(x)f_{\alpha_2}(y)}
\end{equation}

We modify standard replica exchange in two ways: (i) for exact inference, states which violate the constraint are rejected, and (ii)
unlike conventional replica exchange which draws samples only from the zero-temperature chain, we accept states from any chain so long as $f_{\alpha_i}(x) = 1$.

Replica exchange has a number of hyper-parameters: the number of parallel chains, the corresponding temperatures, the swapping schedule.
Several good practices are outlined in \cite{earl2005parallel}.  In practice, we logarithmically space $\alpha$ between a lower and upper bound (e.g., $\log(\alpha_1) = 10^{-5}$, $\log(\alpha_M) = 10^5$), and swap states of chains that are adjacent in temperature ($\alpha_1$ with $\alpha_2$, $\alpha_2$ with $\alpha_3$, etc) periodically.

\section{Implementation}\label{implement}

In this section we describe a generic, lightweight implementation of predicate exchange.
Our approach closely mirrors \citep{wingate2011lightweight, milch20071} in the sense that it provides a language independent layer that can be implemented on top of existing programming languages and modeling formalisms.
Our objective is to twofold: (i) to compute the prior term $p$, approximate likelihood term $\softv{\lk}$, and approximate posterior term $f$ (Equation \ref{approxposterior}) from an arbitrary program $\pi$, and (ii) to perform Replica Exchange MCMC to sample from this posterior.

A program $\pi$ can be an arbitrary composition of deterministic and stochastic procedures, but all stochastic elements must come from a set of known \emph{elementary random primitives}, or ERPs.
ERPs correspond to primitive parametric distribution families, such as the uniform or normal distribution.
Let $\mathcal{T}$ be a set of ERP types.
Each type $\tau \in \mathcal{T}$ must support (i) evaluation of the conditional density $p_\tau(x \mid \theta_1, ..., \theta_n)$, and (ii) sampling from the distribution.
Concretely, a conditioned program $\pi$ is a any nullary program that contains the statements:

\begin{enumerate}
  \item $\textrm{rand}(\tau, n, \theta_1, ...,\theta_n)$ returns a random sample from $p_\tau(x \mid \theta_1, ..., \theta_n)$.  $n$ is a unique named described below.
  \item $\cond(y)$ conditions $\pi$.  It throws an error if $y \in \{0, 1\}$ is 0, and otherwise allows simulation to resume with no effect.
\end{enumerate}

Example Program 1 illustrates a simple conditioned model.

\subsection{Tracked Soft Execution}
The prior term $p$ is computed automatically as the product of random choices in the program. 
That is, let $\pi_{k \mid x_1, ..., x_{k-1}}$ be the k'th ERP encountered in while executing $\pi$, $x_k$ be the value it takes, and $x$ denote the set of all values of all ERPs constructed in the simulation of $\pi$, $p(x)$ is the product:
\begin{equation}\label{productprob}
p(x) = \prod_{k=1}^K p_\tau(x_k \mid \theta_1,..., \theta_n )
\end{equation}
Crucially, the parameters $\theta_1,..,\theta_n$ for each random variable may be fixed values or depend on values of other random variables in $\pi$.

\begin{exprogram}[tb]
\caption{}
\label{prog:ex1}
\begin{algorithmic}
\STATE $x = \textrm{rand}(\mathcal{N}, x, 0, 1)$
\STATE $y = \textrm{rand}(\mathcal{N}, y, 0, 1)$
\STATE $\cond(x > y)$
\STATE {\bfseries Return:} $(x, y)$
\end{algorithmic}
\end{exprogram}

Predicate exchange relies on $\textrm{softexecute}$
(Algorithm \ref{alg:softexecute}), which formalizes the soft execution of a program $\pi$ at temperature $\alpha$, in the context of dictionary $\mathbb{D}$.
$\mathbb{D}$ is a mutable mapping from a set of names to values.
In the context of a particular dictionary, the simulation of a program is deterministic.
This allows the simulation of $\pi$ to be modulated by controlling the elements of $\mathbb{D}$.

$\textrm{softexecute}$ simulates $\pi$ but within a context where (i) variables $\lk_\mathbb{D}$ and $p_\mathbb{D}$ accumulate prior and approximate posterior values, and (ii) the following operators are redefined:

\begin{enumerate}
  \item $\textrm{rand}(\tau, n, \theta_1, ...\theta_n)$ returns $\mathbb{D}(n)$, and in compliance with Equation \ref{productprob} updates $p_\mathbb{D}$ with the conditional density. If $n$ is not a key in $\mathbb{D}$, the distribution is sampled from and $\mathbb{D}(n)$ is updated with this value.  
  \item $a \text{ op } b$ and $\textrm{op } a$ for $\textrm{op} \in \{>, <, =, \land, \lor, \neg\}$ are replaced with the softened counter-parts $\soft{\textrm{ op }} \in \{\soft{>}, \soft{<}, \soft{=}, \soft{\land}, \soft{\lor}, \soft{\neg}\}$.
  \item $\cond(y)$ updates $\softv{\lk}_\mathbb{D}$ with $\softv{\lk}_\mathbb{D} \soft{\land} y$. $y \in [0,1]$ due to soft primitive operators.  
\end{enumerate}

$\textrm{softexecute}$ returns a real value for the approximate posterior of $f$ as a function of the dictionary $\mathbb{D}$.

\paragraph{Control Flow}
Programs may have control flow constructs, such as if-then-else statements.
These may cause $\textrm{softexecute}$ to return a value that is significantly less than $\soft{\lk}_{\inf}$.
This is because if a branch condition is a function of an uncertain value, then several unexplored alternative paths could produce values that are closer to the constraint set.
$\textrm{softexecute}$ is ignorant of thees other possibilities
For illustration, consider Example Program 2 \ref{prog:ex2}.
If $x = -1$ the condition fails, and the predicate relaxation will yield $x \soft{=} -100$, which is significantly larger than if the true branch were taken.

\begin{exprogram}[tb]
\caption{}
\label{prog:ex2}
\begin{algorithmic}
\STATE $x = \textrm{rand}(\mathcal{N}, x, 0, 1)$
\IF {$x > 0$}
\STATE $\cond(x = 1)$
\ELSE
\STATE $\cond(x = -100)$
\ENDIF
\STATE {\bfseries Return:} $x$
\end{algorithmic}
\end{exprogram}

Problems of this form appear in all forms of program analysis.
This problem is called the path explosion problem, since the number of possible paths often increases combinatorially with program size and runtime length.
Automated program testing, which is concerned with finding program paths that yield to failure has developed various strategies \cite{cadar2008exe, sen2005cute}.
% Broadly, these trace the execution of the program and derive in symbol form the branch constraints.
% These constraints are solved to force the execution into branches of the program that incur error states.
% These tools have been scaled to very large problems, in complex real world code.
% Two methods that use heuristics toguide path exploration are [4] (which attempts to explore paths that hit less-often executed statements.
Unlike automated testing, probabilistic inference has the stricter requirement of adhering to the true posterior distribution.
However, in predicate exchange, we have a latitude on all nonunitary values.
This opens up the potential for extending program analysis methods to the probabilistic domain in future work.

\begin{algorithm}[tb]
  \caption{Soft Execution: $\textrm{softexecute}(\pi, \alpha, \mathbb{D})$}
  \label{alg:softexecute}
\begin{algorithmic}
\STATE {\bfseries Input:} program $\pi$, temperature $\alpha$, dictionary $\mathbb{D}$
\STATE Initialize $\softv{\lk}_\mathbb{D} = 1, p_\mathbb{D} = 1$
\STATE Simulate $\pi$ with following subroutines redefined as:   
\ALOOP {$\textrm{rand}(\tau, n, \theta_1, ..., \theta_n)$}
   \IF{$n \in \mathbb{D}$}
   \STATE $x = \mathbb{D}(n)$
 \ELSE
   \STATE $x = $ sample from $p_\tau(x \mid \theta_1, ..., \theta_n)$
   \STATE Update dictionary: $\mathbb{D}(n) = x$
 \ENDIF
 \STATE $p_\mathbb{D} = p_\mathbb{D} \cdot p_\tau(x \mid \theta_1, ..., \theta_m)$
 \STATE Return from subroutine: $x$
\ENDALOOP
\STATE
\ALOOP {$\cond(\lk')$}
  \STATE $\softv{\lk}_\mathbb{D} = \softv{\lk}_\mathbb{D} \cdot \softv{\lk}_\mathbb{D}'$
\ENDALOOP
\STATE
\ALOOP {$\textrm{op}(x, \dots)$ for $\textrm{op} \in \{>, <, =, \land, \lor, \neg\}$}
  \STATE Return from subroutine: $\soft{\textrm{op}}(x, \dots)$ 
\ENDALOOP
\STATE
% \IF{$s = \textrm{rand}(\tau, n, \theta_1, ..., \theta_n)$}
%  \IF{$n \in \mathbb{D}$}
%    \STATE $x = \mathbb{D}(n)$
%  \ELSE
%    \STATE $x = $ sample from $p_\tau(x \mid \theta_1, ..., \theta_n)$
%    \STATE Update dictionary: $\mathbb{D}(n) = x$
%  \ENDIF
%  \STATE $p_\mathbb{D} = p_\mathbb{D} \cdot p_\tau(x \mid \theta_1, ..., \theta_m)$
%  \ELSIF{$s = \cond(\lk')$}
%    \STATE $\lk_\mathbb{D} = \lk_\mathbb{D} \cdot \lk_\mathbb{D}'$
%  \ENDIF
\STATE {\bfseries Return:} $p_\mathbb{D} \cdot \softv{\lk}_\mathbb{D}$
%    \ENDFOR
%    \UNTIL{$noChange$ is $true$}
\end{algorithmic}
\end{algorithm}

\subsection{Replica Exchange}

Predicate exchange (Algorithm \ref{alg:predexchange}) performs replica exchange using $\textrm{softexectute}$ as an approximate posterior.
It takes as input an mcmc algorithm, which simulates an Markov Chain by manipulating elements of the $\mathbb{D}$.
In our experiments, for finite dimensional continuous models we use the No U-Turn Sampler \cite{hoffman2014no}, a variant of Hamiltonion Monte Carlo.
We use reverse-mode automatic differentiation \cite{griewank2008evaluating} to compute the negative log gradient of $f$.
For other models we use standard Metropolis Hastings by defining proposals on elements in the dictionary.
In particular we use the single site MH \cite{wingate2011lightweight} which modifies a single random variable at a time.

% Rep
% Each dictionary should contains all the information required to access values of variables of interest, either explicitly as values in the dictionary, or derivable with the simulator $\pi$. 

\begin{algorithm}[tb]
  \caption{Predicate Exchange}
  \label{alg:predexchange}
\begin{algorithmic}
\STATE {\bfseries Input:} program $\pi$, temperatures $\alpha_1, ...,\alpha_m$, nsamples $n$
\STATE {\bfseries Input:} mcmc, nsamples between swaps $q$ 
\STATE Initialize $\mathcal{D} = $ empty collection of dictionarys
\STATE Initialize $\mathbb{D}^{\textrm{init}}_1,...,\mathbb{D}^{\textrm{init}}_m$ empty dictionarys
\STATE Define $f_{\alpha_i}(\mathbb{D}) = \textrm{softexecute}(\pi, \alpha_i, \mathbb{D})$
\REPEAT
  \FOR{$i=1$ {\bfseries to} $m$}
    \STATE { $\mathbb{D}_1,...,\mathbb{D}_q = $ $q$ mcmc samples at temp $\alpha_i$}, from $\mathbb{D}^{\textrm{init}}_i$
    \STATE $\mathbb{D}^{\textrm{init}}_i = \mathbb{D}_q$
    \FOR{$j=1$ {\bfseries to} $q$}
      \IF {$f_{\alpha_1}(\mathbb{D}_j) = 1$}
        \STATE append $\mathbb{D}_j$ to $\mathcal{D}$
      \ENDIF
    \ENDFOR
  \ENDFOR
  \FOR{$i = m$ {\bfseries down to} $2$}
    \STATE $j = i - 1$
    \STATE $p = {f_{\alpha_i}(\mathbb{D}_j)f_{\alpha_j}(\mathbb{D}_i)}/{f_{\alpha_i}(\mathbb{D}_i)f_{\alpha_j}(\mathbb{D}_j)}$
    \IF{$p >$ random sample in $[0, 1]$}
      \STATE swap $\alpha_i$ with $\alpha_j$
    \ENDIF
  \ENDFOR
\UNTIL{$\mathcal{D}$ has $n$ elements}
\STATE {\bfseries Return:} $\mathcal{D}$
\end{algorithmic}
\end{algorithm}

\section{Experiments}

% \begin{figure}
% 	\centering
% 	\includegraphics[width=0.9\linewidth,natwidth=2000,natheight=2400]{grid3.png}
% 	\caption{Posterior samples on different problems.  RE: replica exchange, SSMH: Single Site Metropolis Hastings, HMC: Hamiltonian Monte Carlo.  (Left) is target density, and following elements are histograms from computed samples.}
% 	\label{fig:grid}
% \end{figure}

\begin{figure}[!htb]
\centering
\includegraphics[width=0.8\linewidth]{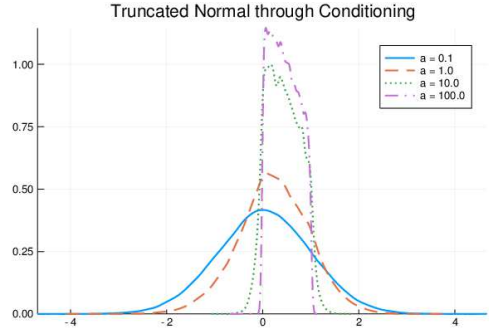}

% \begin{minipage}{0.45\linewidth}
% \includegraphics[width=\linewidth]{truncated}
% \end{minipage}%
% \begin{minipage}{0.45\linewidth}
% 	%\includegraphics[width=.16\linewidth, trim={1.7cm, 1.6cm, 1.3cm, 1.5cm}, clip]{0-1}
% 	\includegraphics[width=.45\linewidth, trim={1.7cm, 1.6cm, 1.3cm, 1.5cm}, clip]{1-0}
% 	\includegraphics[width=.45\linewidth, trim={1.7cm, 1.6cm, 1.3cm, 1.5cm}, clip]{10-0}
	
% 	\includegraphics[width=.45\linewidth, trim={1.7cm, 1.6cm, 1.3cm, 1.5cm}, clip]{100-0}
% 	\includegraphics[width=.45\linewidth, trim={1.7cm, 1.6cm, 1.3cm, 1.5cm}, clip]{1000-0}				
	
% %	\fbox{\includegraphics[width=.16\linewidth, trim={1.7cm, 1.6cm, 1.3cm, 1.5cm}, clip]{0-1}}
% %	\fbox{\includegraphics[width=.16\linewidth, trim={1.7cm, 1.6cm, 1.3cm, 1.5cm}, clip]{1-0}}
% %	\fbox{\includegraphics[width=.16\linewidth, trim={1.7cm, 1.6cm, 1.3cm, 1.5cm}, clip]{10-0}}
% %	\fbox{\includegraphics[width=.16\linewidth, trim={1.7cm, 1.6cm, 1.3cm, 1.5cm}, clip]{100-0}}
% %	\fbox{\includegraphics[width=.16\linewidth, trim={1.7cm, 1.6cm, 1.3cm, 1.5cm}, clip]{1000-0}}				
% 	\end{minipage}
	\caption{Left: Density from samples of Gaussian truncated to $[0, 1]$ through conditioning. Right: Conditioning on $X = Y$ where $X$ and $Y$ are independent normal distributions; shown at different temperatures.}
	\label{fig:density}
\end{figure}

\paragraph{Small Models}
In Figure \ref{fig:density} we demonstrate two examples of conditioning on predicates which are non trivial.
First we show that the conditioning can be used to truncate a Gaussian distribution, and the approximation behavior at varying temperatures.  Second we show that two independent random variables can be made equal.  While simple, both are a challenge for probabilistic programming systems because they prevent automatic calculation of the likelihood.

\paragraph{Glucose Model}
Type 2 diabetes is a prevalent and costly condition.
Keeping blood glucose within normal limits helps prevent the
long-term complications of Type 2 diabetes like diabetic neuropathy and diabetic retinopathy \citep{brownlee2006glycemic}. Models to predict the trajectories of blood glucose aid in keeping glucose within
normal limits \citep{zeevi2015personalized}. Traditional models have been built from compositions of differential equations \citep{albers2017personalized,levine2017offline} whose parameters are estimated separately for each patient. An alternative approach would be to use a flexible sequence model like an RNN. The problem with this approach is that an RNN can extrapolate to glucose values incompatible with human physiology. This is especially a problem where we have patients with only a few blood glucose measurements. To build an RNN model that respects physiology, we condition on it.

\begin{figure}[!htb]
	\centering
    %\fbox{\includegraphics[width=0.30\linewidth, trim={1.cm, 0.1cm, 1.3cm, .5cm}, clip]{rnnsamples-no-tie-py}}
	\includegraphics[width=0.45\linewidth, trim={1.cm, 0.1cm, 1.3cm, .5cm}, clip]{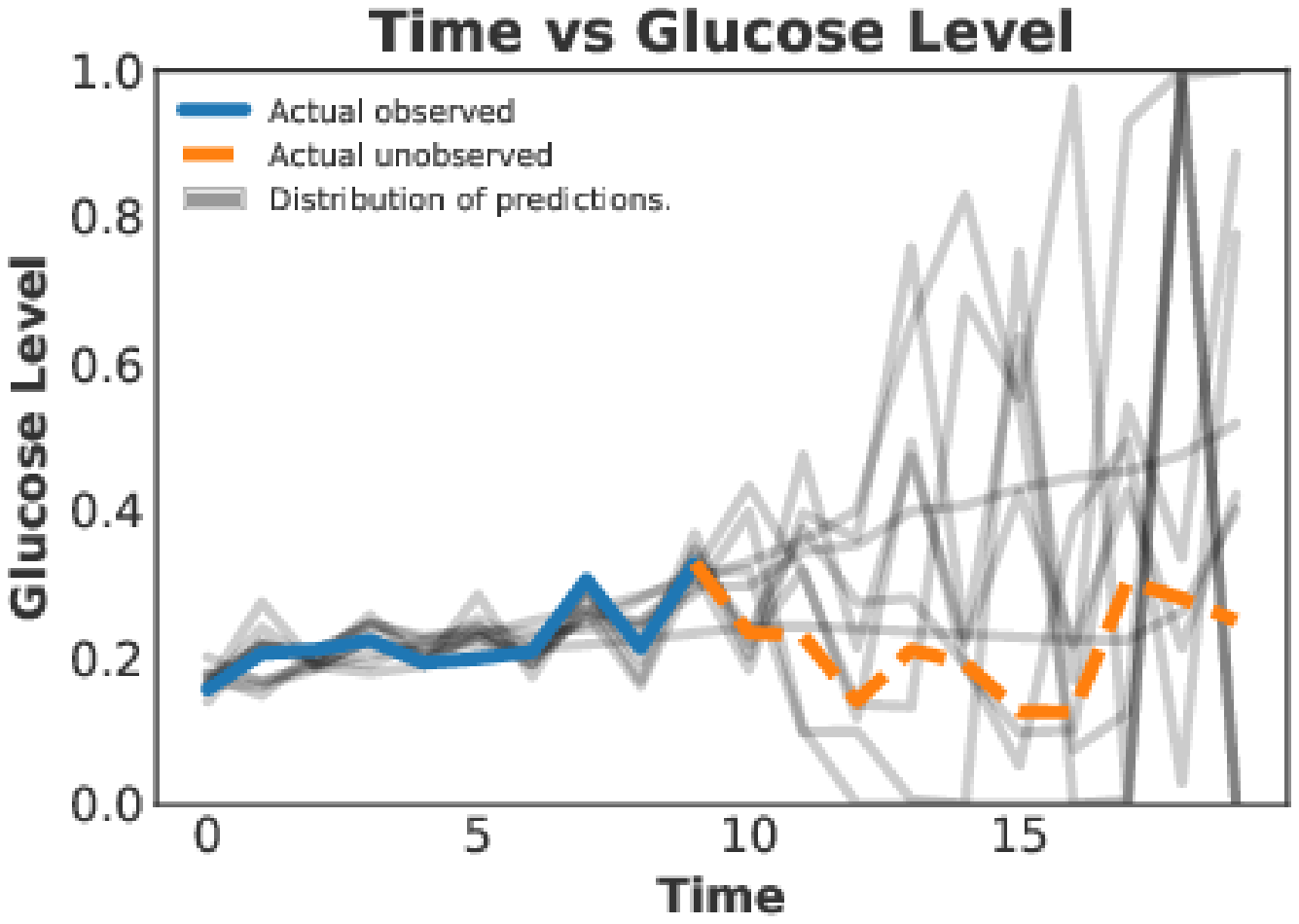}
	\includegraphics[width=.45\linewidth, trim={1.cm, 0.1cm, 1.3cm, .5cm}, clip]{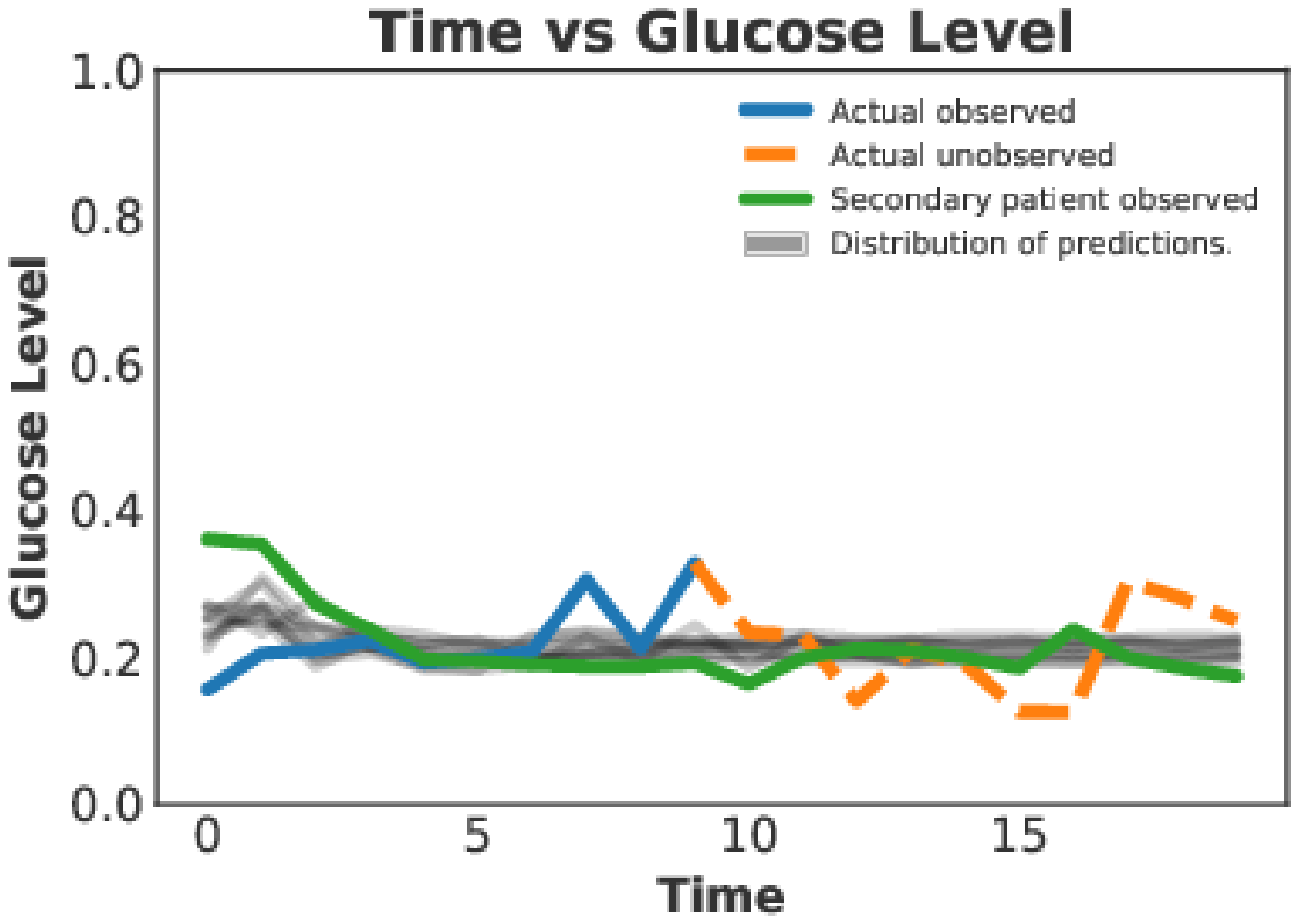}
	%\fbox{\includegraphics[width=0.30\linewidth, trim={1.cm, 0.1cm, 1.3cm, .5cm}, clip]{rnnsamples-py}}
	%\fbox{\includegraphics[width=0.30\linewidth, trim={1.cm, 0.0cm, 1.0cm, .5cm}, clip]{delta_mse}}
	\includegraphics[width=0.9\linewidth, trim={0.7cm, 0.0cm, .7cm, .1cm}, clip]{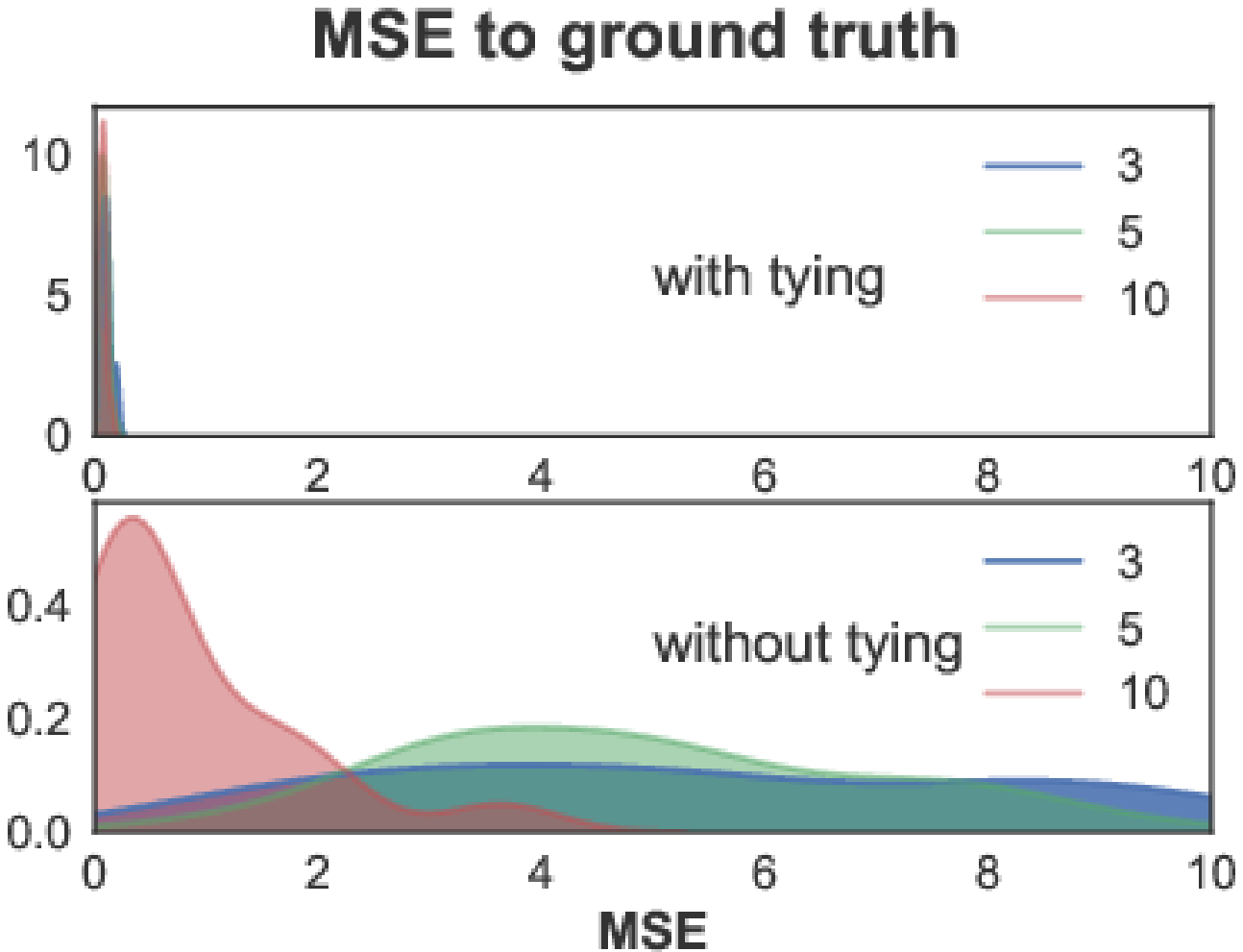}
		
		\caption{Left: Actual (dotted) and predicted trajectories that were learned using a partial trajectory. Center: Distribution of predicted trajectories learned only using the first ten data points and a tie with a secondary patient. Right, top: MSE when tie is present.  Right, bottom: without tie.  Tying expectations has dramatic influence on prediction error, while as more data is observed, the effect of tying decreases.}
		\label{fig:rnn-samples}
\end{figure}

We compare the independent RNN model to the one with declarative knowledge on a second patient from Physionet \citep{moody2001physionet}.
Figure \ref{fig:rnn-samples} plots the results performed on more than 300 pairs of patients.
We see that the conditional model simulates
more realistic glucose dynamics for the patient 
with only a short observed time-series.

\section{Discussion}
In this work we expanded the class of predicates that probabilistic models can be conditioned on in practice.

Problems of this form appear in all forms of program analysis.
This problem is called the path explosion problem, since the number of possible paths often increases combinatorially with program size and runtime length.
Automated program testing, which is concerned with finding program paths that yield to failure has developed various strategies \cite{cadar2008exe, sen2005cute}.
% Broadly, these trace the execution of the program and derive in symbol form the branch constraints.
% These constraints are solved to force the execution into branches of the program that incur error states.
% These tools have been scaled to very large problems, in complex real world code.
% Two methods that use heuristics toguide path exploration are [4] (which attempts to explore paths that hit less-often executed statements.
Unlike automated testing, probabilistic inference has the stricter requirement of adhering to the true posterior distribution.
However, in predicate exchange, we have a latitude on all nonunitary values.
This opens up the potential for extending program analysis methods to the probabilistic domain in future work.

% The objective of this contribution is to expand the class of predicates that probabilistic models can be effectively conditioned on

% Our methodology falls broadly within the domain of approximate bayesian computation, in the sense that we equip spaces with a distance metric.
% At a high level, this can be understood as extracting more information out of a predicate, which unmodified provides only 0 or 1, which is insufficiently informative for any inference procedure to exploit.
% By an large, inference in high-dimensional models has been restricted to fininte dimensional continous models.
% This remains the case for our method, methods discrete models remain a challenge

% Control flow remains a challenging problem, for fundamental reasons.

% There are several inference strategies other than replica exchange MCMC that could be constructed onto of a relaxed predicate.
% Maximum posterior inference is the most immediate option, which would entail maximizing equation \ref{}.

% Black-box inference methods have gained significant traction due to how general, flexible, a vs grey box

% \input{key}

\bibliography{bib.bbl}
\bibliographystyle{icml2018}

\end{document}